\documentclass[lettersize,journal]{IEEEtran}
\usepackage{amsmath,amsfonts}
\usepackage{algorithmic}
\usepackage{algorithm}
\usepackage{array}
\usepackage[caption=false,font=normalsize,labelfont=sf,textfont=sf]{subfig}
\usepackage{textcomp}
\usepackage{stfloats}
\usepackage{url}
\usepackage{verbatim}
\usepackage{graphicx}
\usepackage{cite}
\usepackage[font=small,skip=2pt]{caption} 

\begin{document}
\title{Self-Wearing Adaptive Garments via Soft Robotic Unfurling}

\author{Nam Gyun Kim$^{1}$, William E. Heap$^{2}$, Yimeng Qin$^{2}$, Elvy B. Yao$^{2}$, Jee-Hwan Ryu$^{3*}$, and Allison M. Okamura$^{2}$

\thanks{Manuscript created June 2025. This research was supported in part by National Science Foundation Award 2345769, Stanford University, Korean Institute for Advancement of Technology(KIAT) grant funded by the Korea Government(MOTIE) (RS-2024-00435502, Human Resource Development Program for Industrial Innovation(Global)), and the National Research Foundation of Korea under Grant RS-2025-00554618. (\it{Corresponding author: Jee-Hwan Ryu.})}

\thanks{$^{1}$Nam Gyun Kim is with the Robotics Program, KAIST, 34141, Daejeon, South Korea}%

\thanks{$^{2}$William E. Heap, Yimeng Qin, Elvy B. Yao, and Allison M. Okamura are with the Department of Mechanical Engineering, Stanford University, 94305, California, United States of America}%

\thanks{$^{3}$Jee-Hwan Ryu is with the Department of Civil and Environmental Engineering, KAIST, 34141, Daejeon, South Korea {\tt\small jhryu@kaist.ac.kr}}}



\maketitle

\begin{abstract}
Robotic dressing assistance has the potential to improve the quality of life for individuals with limited mobility. Existing solutions predominantly rely on rigid robotic manipulators, which have challenges in handling deformable garments and ensuring safe physical interaction with the human body. Prior robotic dressing methods require excessive operation times, complex control strategies, and constrained user postures, limiting their practicality and adaptability. This paper proposes a novel soft robotic dressing system, the Self-Wearing Adaptive Garment (SWAG), which uses an unfurling and growth mechanism to facilitate autonomous dressing. Unlike traditional approaches, the SWAG conforms to the human body through an unfurling-based deployment method, eliminating skin-garment friction and enabling a safer and more efficient dressing process. We present the working principles of the SWAG, introduce its design and fabrication, and demonstrate its performance in dressing assistance. The proposed system demonstrates effective garment application across various garment configurations, presenting a promising alternative to conventional robotic dressing assistance.
\end{abstract}

\begin{IEEEkeywords}
self-wearing, autonomous dressing, robotic assistance, soft robotics.
\end{IEEEkeywords}

\section{Introduction}
\IEEEPARstart{D}{ressing} is a fundamental activity of daily living that directly impacts independence and quality of life. For individuals with physical disabilities, the elderly, and those recovering from injuries, dressing can be a significant challenge \cite{mann2005problems}. The inability to dress independently often leads to a loss of autonomy, increased reliance on caregivers, and a diminished sense of dignity. Consequently, the development of effective dressing assistance technologies can enhance accessibility and promote greater inclusivity for individuals with limited dexterity and mobility \cite{bostrom2014functional, vogelpohl1996can}. To address dressing-related difficulties, various technological approaches have been investigated, including robotic manipulators that assist in donning conventional garments \cite{tamei2011reinforcement}.

\begin{figure}[t]
\centerline{\includegraphics[width=85mm]{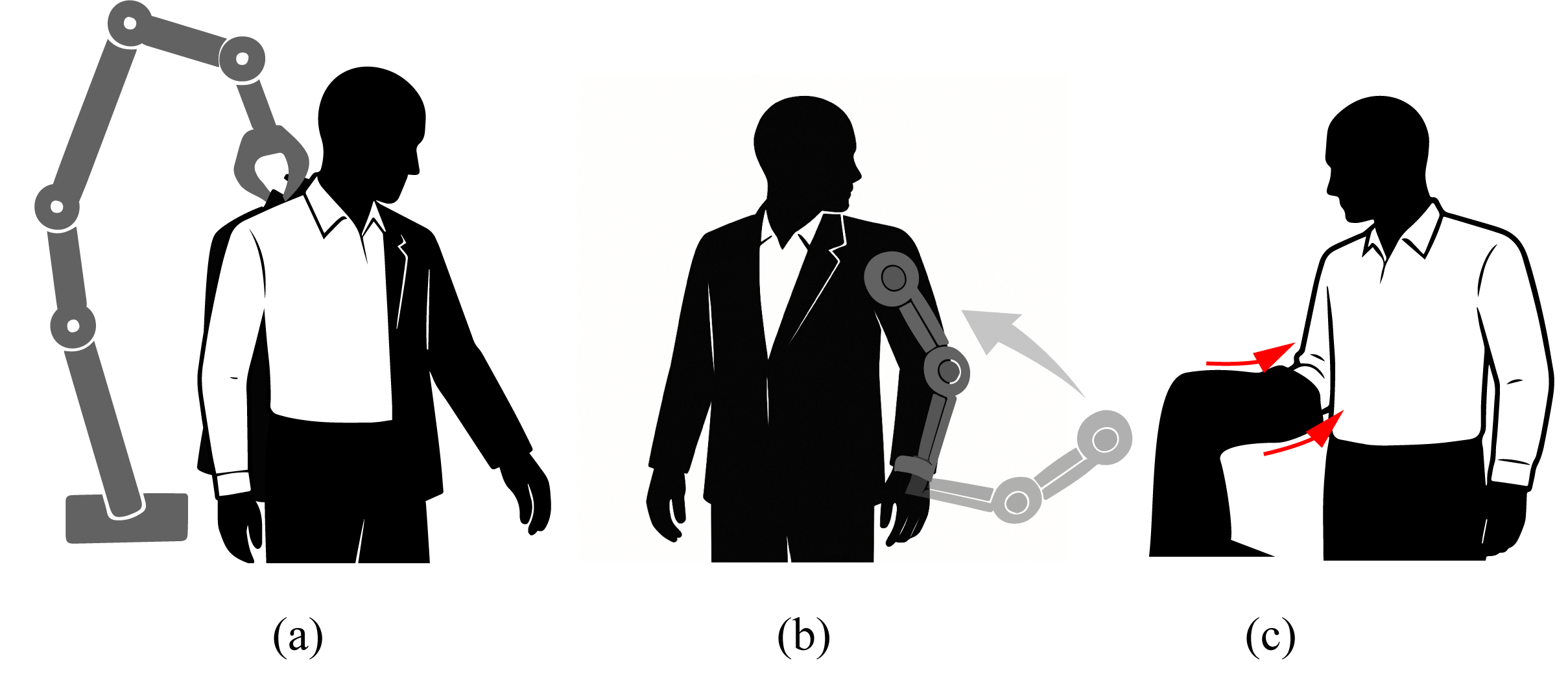}}
\caption{Types of robot-assisted dressing systems. (a) An external robot manipulates conventional garments to assist the user. (b) Specialized wearable assistive/rehabilitative devices are capable of autonomous donning and doffing. (c) Proposed approach: A self-wearing system that augments existing garments with integrated soft robotic mechanisms for autonomous dressing.}
\label{fig:Donningtype}
\end{figure}

Efforts focused on dressing users with commercially available clothing still face significant limitations. Most existing research focuses on algorithms that enable a robotic manipulator to place a sleeve onto a user's arm (Fig.~\ref{fig:Donningtype}(a)). Achieving successful dressing for a single arm is a critical milestone, implying that the same approach can be extended to the other limb \cite{gao2016iterative, erickson2018deep, kapusta2019personalized, zhang2019probabilistic}. However, robotic manipulators must handle highly flexible fabrics \cite{ramisa2014learning} during continuous physical interaction with the human body \cite{colome2015friction}. The usability of these systems commonly suffers from excessive operation times and requires users to maintain fixed postures throughout the dressing process. Moreover, the high cost of sensors, the complexity of control algorithms, and the challenges associated with accurate sensing further hinder their practicality. The rigid structures of these systems also pose safety concerns \cite{delgado2021safety}. Overall, these limitations highlight the need for alternative solutions that offer improved compliance, adaptability, and safety in robot-assisted dressing.

Meanwhile, a rehabilitative device capable of self-donning and doffing has been proposed (Fig.~\ref{fig:Donningtype}(b)). Park et al. developed a self-wearable exoskeleton system for individuals with lower-limb impairments \cite{park2024design}. However, such a device is designed for specific clinical or therapeutic purposes, and its rigid structures render it incompatible with the requirements of general-purpose dressing assistance. 

A promising approach to addressing these challenges is the use of soft robots, which have emerged as an alternative to conventional rigid robots due to their inherent compliance and flexibility \cite{rus2015design}. Unlike traditional robotic systems, soft robots can have continuum structures and conform to complex, unknown objects, making them well-suited for tasks that require delicate manipulation and safe physical interaction with humans \cite{shintake2018soft, polygerinos2017soft}. Their potential for simple fabrication and low materials costs further enhance their practicality while addressing many of the limitations associated with rigid robots \cite{zongxing2020research}.

The effectiveness of soft robots in assistive and human-interactive applications has been demonstrated in various studies. Soft robotic arms have been widely explored for various assistive tasks, leveraging their inherent compliance to enable safe and adaptive interaction with the human body \cite{chen2022review}. Soft robotic gloves and exosuits have also been explored for assisting human movement, demonstrating significant potential in improving mobility and facilitating rehabilitation \cite{polygerinos2015soft, jiralerspong2018novel}. These studies underscore the advantages of soft robotics in applications where adaptability and safety for human-robot interaction are paramount.

Despite these advantages, the application of soft robotics in dressing assistance remains largely unexplored. Kim et al. \cite{kim2022knitskin} proposed an inchworm-like soft robotic locomotion system that autonomously pulls a sleeve onto a human arm, while Agharese et al. \cite{agharese2018hapwrap} developed a growing, self-wrapping haptic device capable of conforming to various forearm sizes. However, neither approach addresses the broader challenge of dressing with general garments. Given that dressing tasks could inherently benefit from the compliance and adaptability of soft robots, which enables safer and more natural interactions than rigid systems, it is notable that no soft robotic solution has yet been proposed for general-purpose garment dressing.

In this paper, we introduce a novel soft robotic dressing assistance system: the Self-Wearing Adaptive Garment (SWAG). This system augments existing garments by integrating a soft robotic donning mechanism (Fig.~\ref{fig:Donningtype}(c)). The SWAG employs an unfurling and growth mechanism to autonomously don clothing while conforming to the human body. Using tip unfurling, the SWAG minimizes friction between the garment and the skin, enabling a safer and more adaptive dressing process. Furthermore, this approach significantly reduces the overall dressing time compared to conventional robotic arm-based methods. Whereas existing rigid robotic solutions often treat the human body as an obstacle, requiring complex trajectory planning to avoid collisions during the donning process, the proposed soft robotic self-wearing technology leverages its compliance to adapt to the user's body. In doing so, the human body itself implicitly guides the donning trajectory, allowing the garment to naturally follow the contours of the body, even when the user's posture varies during the process. This approach not only enhances safety and adaptability but also simplifies sensing requirements and reduces the need for computationally intensive control processes such as vision-based motion planning during dressing. We present the working principles of SWAGs, along with design guidelines, prototype fabrication process, and experimental validation to characterize its capabilities. Furthermore, its effectiveness is demonstrated through a human subject demonstration.

The remainder of this paper is organized as follows: Section II describes the proposed system, while Section III details its operational principles. Section IV outlines the prototype fabrication process, and Section V presents experimental results and demonstrations. Finally, Section VI discusses the findings and explores future research directions.

\section{Proposed Mechanism}
When removing tight clothing, it is often easier to turn the garment inside out while pulling it off. This ease of removal stems from the garment detaching without sliding against the skin. The proposed robot in this paper leverages this characteristic to assist with dressing. The proposed self-wearing adaptive garment (SWAG) employs an inverting, unfurling motion to cover non-uniform objects in a stable and adaptive manner. To facilitate minimal friction between the fabric and the object while enclosing the object, a SWAG transports the outer membrane to the tip and unfurls it over the object to achieve complete coverage.

\begin{figure}[t]
\centerline{\includegraphics[width=85mm]{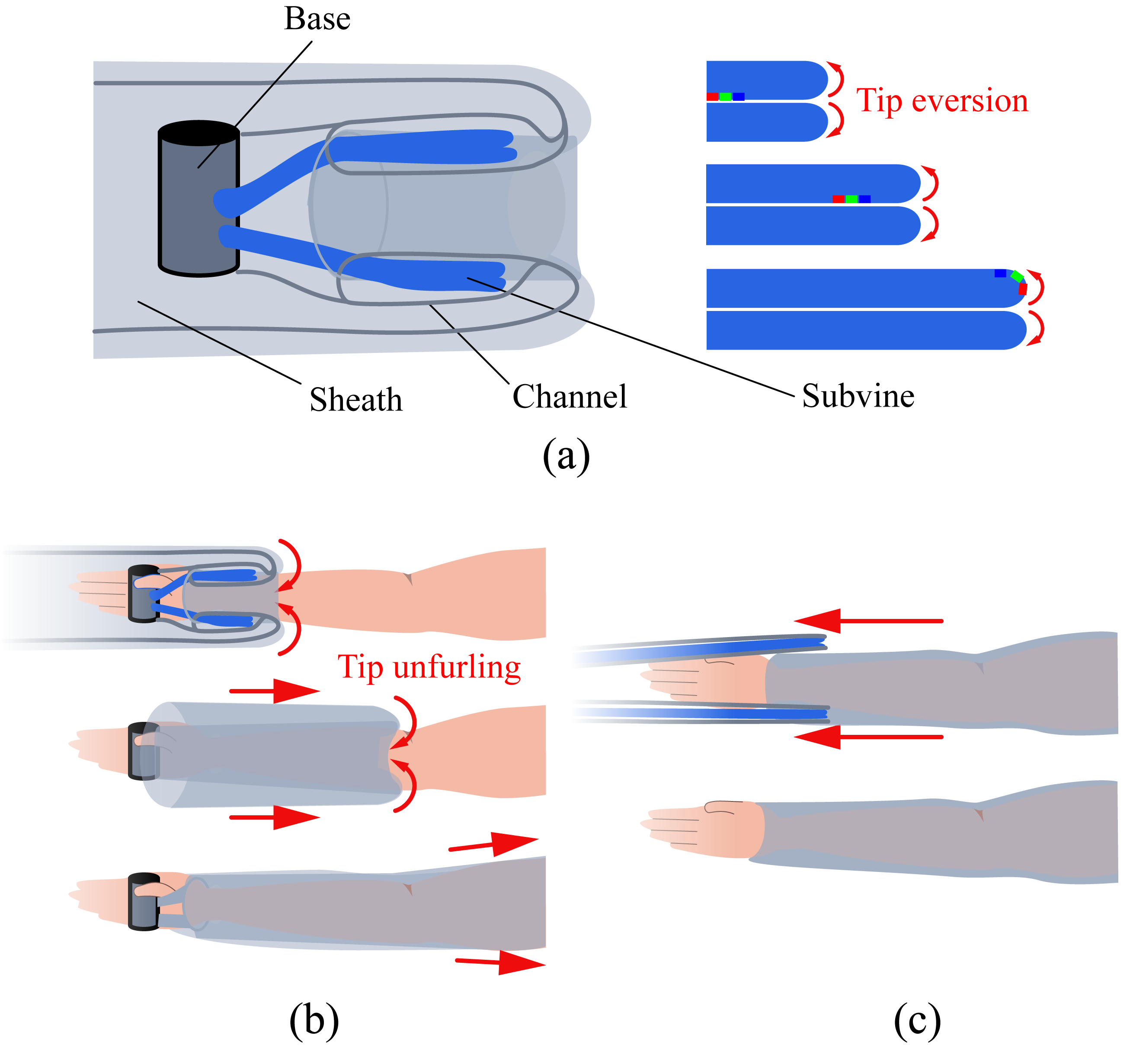}}
\caption{Concept of the unfurling-based donning mechanism: (a) Schematic illustration of the proposed system. A subvine grows within a channel integrated into the garment sheath. (b) Donning process. As the subvine grows forward, it pushes the distal end of the sheath, inducing an unfurling motion that enables autonomous garment donning. (c) Doffing process. The subvine retracts, allowing the sheath to be easily withdrawn from the body.}
\label{fig:Concept}
\end{figure}

Fig.~\ref{fig:Concept} illustrates the fundamental working principle of the proposed approach. As shown in the figure, the mechanism functions by unfurling a thin sheath to wrap around an object. The sheath employed in this mechanism is a thin fabric formed into a cylinder, along which small-diameter channels are distributed circumferentially. Along these channels, an eversion mechanism, referred to as a subvine, extends inside. By leveraging the principle of soft growing robots, known as ``vine robots'', which continuously extend from their tip while minimizing friction with their surroundings \cite{hawkes2017soft}, the subvine generates a force that propels the sheath forward. As a result, the sheath unfurls and envelops the object (i.e., the human body) with minimal friction. A SWAG can function either as a garment itself or as a mechanism that assists in donning other garments. If an additional layer, such as clothing or a covering material, is placed over the sheath, the added layer can be applied to the object seamlessly without friction. Once the operation is complete, the overall mechanism can be retracted simply by pulling it back, leaving only the covering material on the object.

This unfurling of the sheath can be regarded as the inverse operation of a vine robot. However, unlike conventional vine robots, where the outer membrane remains stationary while the tip inverts during retraction, this approach involves active movement of the outer membrane to achieve unfurling. Similar mechanisms have been demonstrated, where vine robots were utilized as grippers \cite{li2020bioinspired, root2021bio, sui2022bioinspired}. However, their approach required full enclosure and continuous connectivity to sustain internal pressure within the membrane, which limits its applicability to dressing assistance. 

In this paper, we employ two design strategies that enable this unfurling behavior without pressurizing the entire membrane, making it more suitable for dressing assistance. Similar to the retraction method of the soft growing robot proposed by Kim et al., our approach also utilizes a method that increases the stiffness of the inner membrane while simultaneously generating an outward force to naturally transition the outer membrane into the inner membrane \cite{kim2023self}. To achieve greater stability, higher force generation, and applicability under static conditions, the proposed approach employs the subvine, which completely isolates the internal pressure of the channel within the inner membrane from that of the outer membrane. This method, as demonstrated in Seo et al.\cite{seo2024inflatable}, enables a stable lengthening operation while avoiding inflation of the whole structure. Such actuation allows the sheath to operate along the body without requiring inflation, thereby maintaining high compliance and eliminating any unnecessary squeezing force applied to objects or human skin. This strategy of excluding inflation of the entire sheath is important, as simply inflating the entire sheath while maintaining tension in the inner membrane is ineffective for inducing unfurling. The slack outer fabric tends to expand and kink under pressure, similar to buckling behavior in conventional vine robots, rather than transmitting force to the distal tip to initiate unfurling. As a result, treating the entire sheath as a single vine-like structure fails to achieve reliable unfurling propagation. To address this limitation, sub-vine-based strategies that enable localized and controllable unfurling were a crucial design choice, making the system more suitable for dressing assistance applications.

\section{Working Principle}\label{section:working principle}

\begin{figure}[t]
\centerline{\includegraphics[width=80mm]{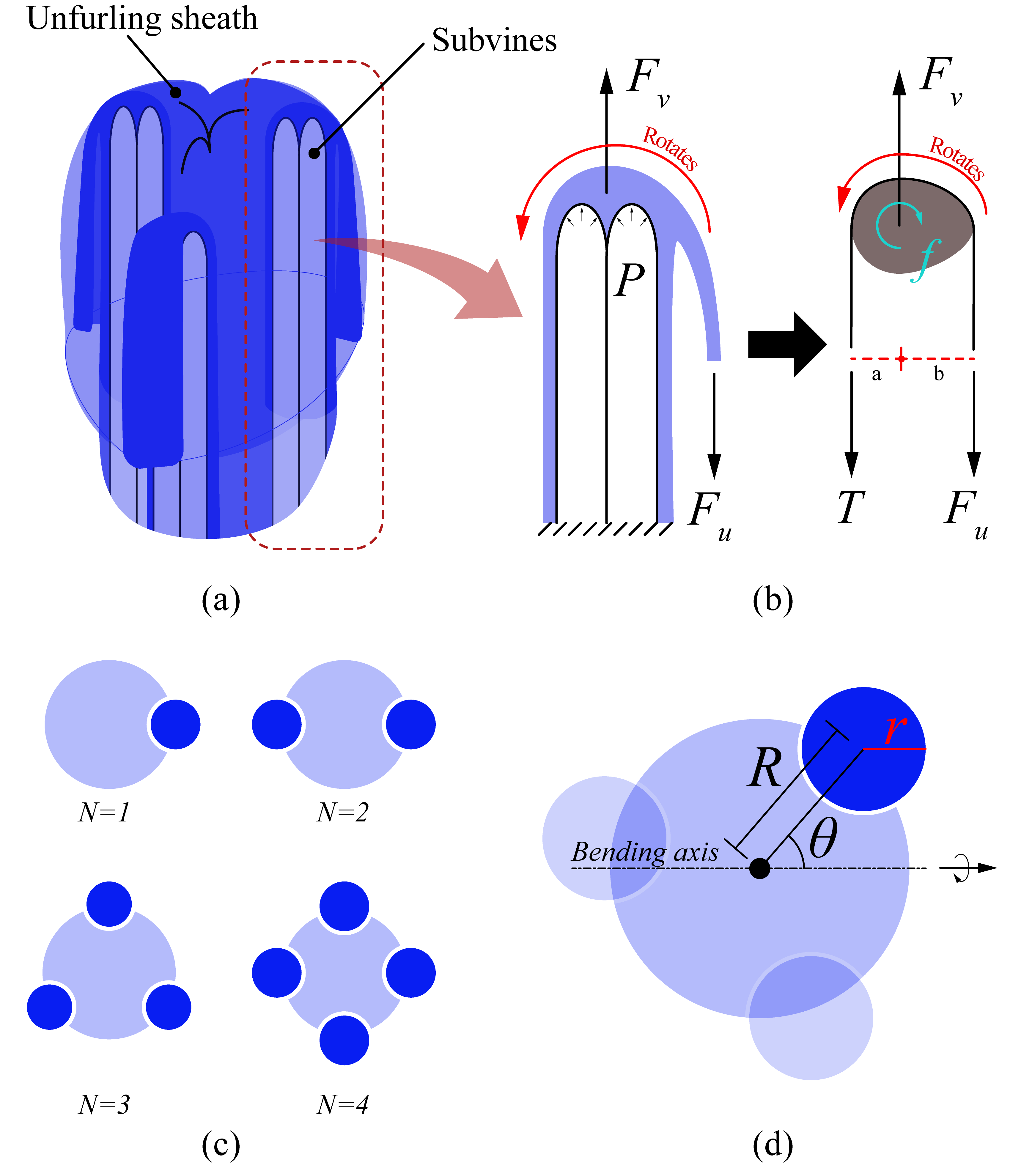}}
\caption{Principle of the proposed unfurling mechanism: (a) Conceptual illustration of the proposed mechanism. (b) Schematic diagram of the force analysis for a single subvine. The unfurling force generated by each subvine is modeled as a nonlinear frictional pulley system. (c) Cross-sectional views of the mechanism with varying numbers of subvines. (d) Geometric parameters used for calculating the area moment of inertia.}
\label{fig:Principle}
\end{figure}

Figure~\ref{fig:Principle} illustrates the free-body diagram representing the unfurling mechanism of the SWAG. The interaction between the tip of a subvine and the tip of a channel, which slide against each other, can be modeled as a frictional, non-linear pulley mechanism. For each channel, the free-body diagram can be expressed as shown in Fig.~\ref{fig:Principle}(b), where $F_v$ represents the force induced by the growth of the subvine, $T$ denotes the reaction tension applied to the inner side of the sheath, and $F_u$ is the unfurling force generated by a single subvine.

The torque equilibrium at the pulley can be expressed as:

\begin{equation}\label{eq:torque_equilibrium}
    F_v \cdot a = F_u \cdot (a+b) + f,
\end{equation}
where $a$ and $b$ represent the distances to the points of force application, and $f$ is a residual term accounting for friction and material deformation. The force generated by a subvine is expressed as:

\begin{equation}\label{eq:vine_force}
    F_v = PA,
\end{equation}
where $P$ represents the internal pressure of the subvine, and $A$ is the cross-sectional area of the subvine. Thus, the final equilibrium equation is:
\begin{equation}\label{eq:final_pressure_eq}
 F_u = \left(\frac{a}{a+b}\right)\cdot (PA - f).   
\end{equation}

If the number of subvines is $N$, and we consider lifting a cloth against gravity, the unfurling force required for each subvine is given by:
\begin{equation}
F_u = \frac{Mg}{N},
\end{equation}
where $M$ represents the mass of the cloth, and $g$ is the gravitational acceleration. Thus, the final equation can be expressed as:

\begin{equation}\label{eq:final_eq}
 Mg = N \left(\frac{a}{a+b}\right) \cdot (PA - f).
\end{equation}

From this equation, it can be observed that as the number of subvines increases, the weight of the fabric or garment to be lifted is distributed among multiple subvines. Consequently, the required pressure per subvine decreases while still enabling the system to operate effectively.

\begin{figure}[t]
\centerline{\includegraphics[width=95mm]{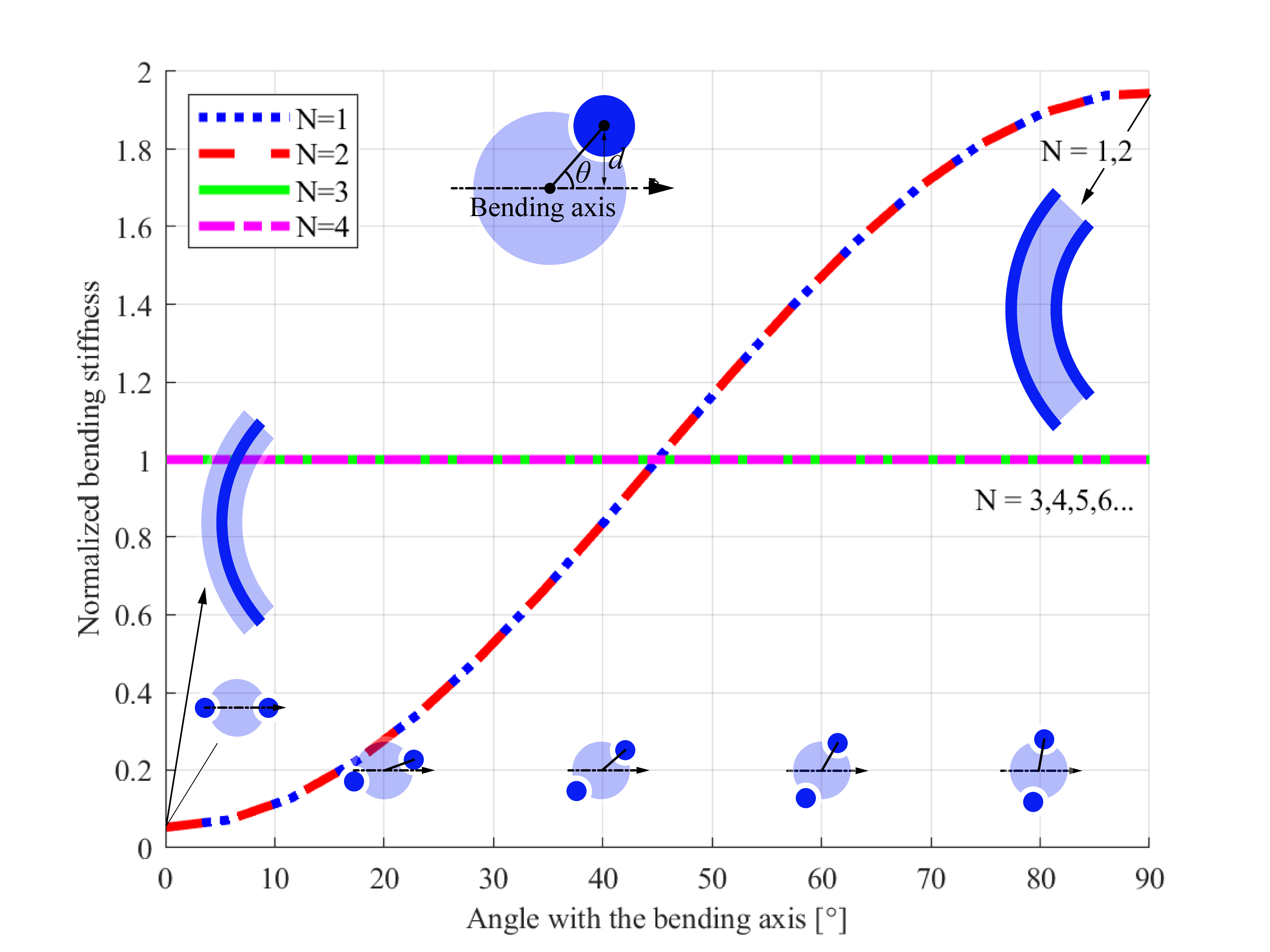}}
\caption{Normalized bending stiffness as a function of the angle between a subvine and the bending axis. Systems with varying numbers of subvines are shown (N = 1,2,3,4...), with the subvine pressure modified to keep each system's propulsive force constant.}
\label{fig:simulation}
\end{figure}

However, increasing the number of subvines also affects the compliance of the system. For example, in Fig. \ref{fig:Principle}(c), where the subvines are arranged in an axisymmetric manner, the increased area moment of inertia may reduce compliance despite the lower pressure. In Fig. \ref{fig:Principle}(d), the subvines positioned at an offset from the bending axis contribute to an increase in the bending stiffness of the overall structure. The increase in area moment of inertia depending on the distance between a single subvine and the bending axis is:

\begin{equation}\label{eq:moment_of_inertia}
    I_{x'} = I_x + A d^2,
\end{equation}
where $I_{x'}$ is the enlarged area moment of inertia of a subvine relative to the bending axis, $I_x$ is the area moment of inertia of a subvine, $A$ is the cross-sectional area, and $d$ represents the shortest distance between the bending axis and the centroid.

Because subvines are cylindrical in shape, their area moment of inertia and cross-sectional area can be defined as:

\begin{equation}\label{eq:cylinder_moment}
    I_x = \frac{\pi D^4}{64}, \quad A = \frac{\pi D^2}{4},
\end{equation}
where $D$ is the diameter of the subvine.

These equations indicate that as subvines are positioned further from the bending axis, their contribution to bending stiffness increases, thereby reducing the overall compliance of the system. Fig.~\ref{fig:simulation} illustrates the normalized bending stiffness of the entire structure with varying numbers of axisymmetrically placed subvines, with subvine pressures set so that every configuration generates the same propulsion force. When the number of subvines is one or two, as depicted in Fig.~\ref{fig:Principle}(d), the bending stiffness varies with the bending axis angle. However, when the number of subvines is three or more, symmetry results in the bending stiffness remaining uniform across all bending axis angles.

From these observations, the following design guidelines can be inferred. If the major bending axis can be predetermined during the design phase, or if it remains close to the axis that minimizes bending stiffness in most scenarios, it is preferable to keep the number of subvines at two or fewer. This consideration is particularly crucial when the SWAG must operate along curved trajectories, where compliance plays a significant role.

Beyond this, the number of subvines should be determined based on the material's tensile strength, ensuring that the selected configuration can withstand the required internal pressure. Additionally, while the current modeling assumes a linear force distribution for simplicity, in real-world scenarios, an increased number of subvines introduces additional interference. This leads to an increase in the nonlinear term $f$, thereby requiring higher actual pressures. Consequently, maintaining a reasonably low number of subvines is recommended to balance operational efficiency and structural compliance.

\section{Prototype Fabrication}
The SWAG consists of two main components: the sheath, which performs the unfurling, and the subvine, which generates the unfurling force. In its simplest form, a SWAG can be constructed by placing a preloaded subvine inside the sheath. In this paper, both the sheath and the subvine are fabricated using thermoplastic polyurethane (TPU)-coated ripstop nylon. To reduce the bending stiffness of the inflated cylindrical subvine while promoting axial elongation of the sheath, both components are fabricated such that the braided direction of the ripstop forms a 45-degree angle with the axial direction. This fabrication approach ensures low stiffness, resulting in reduced restoring bending moment and increased compliance during deflection.

\begin{figure}[t]
\centerline{\includegraphics[width=80mm]{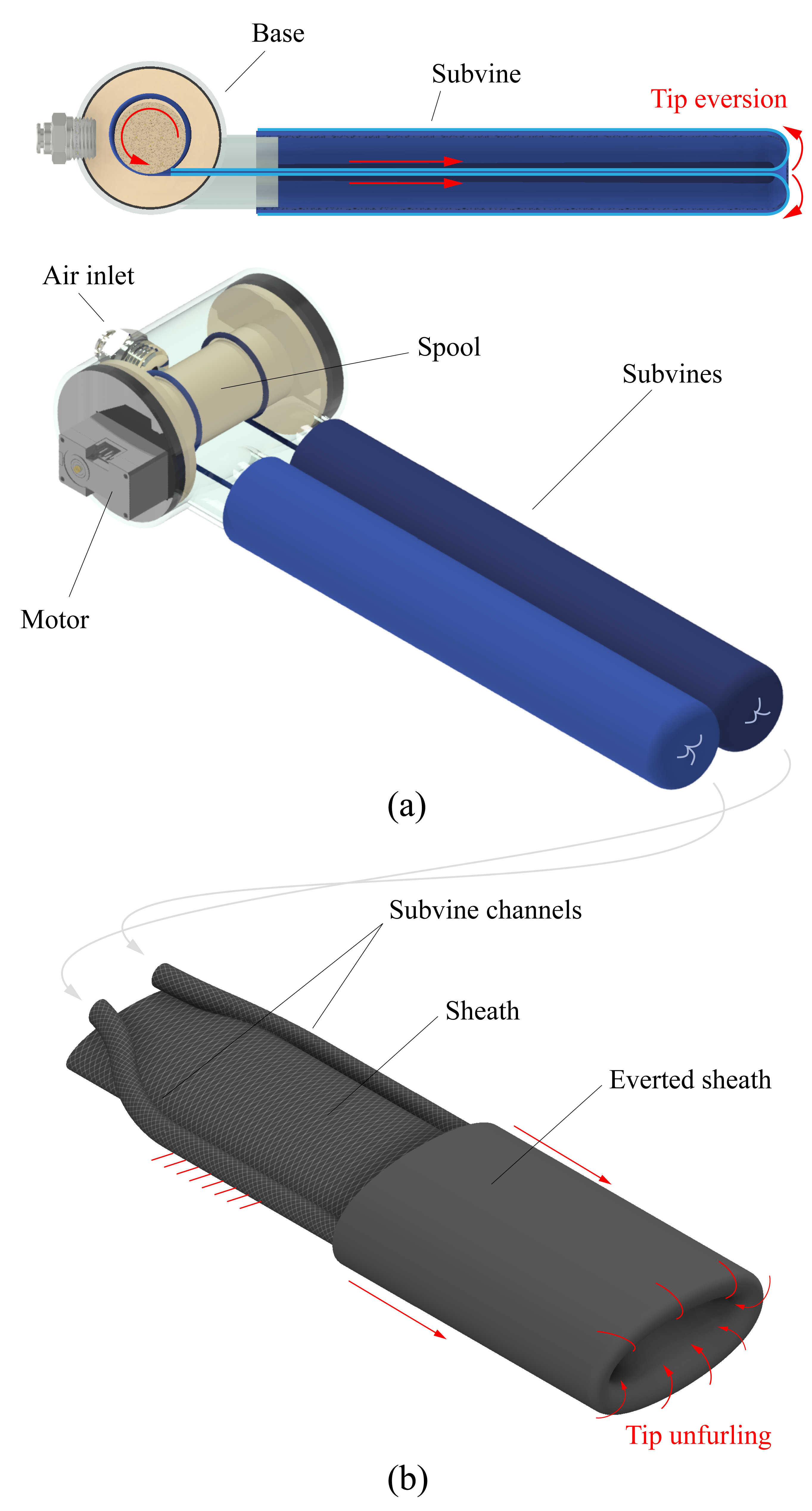}}
\caption{Hardware implementation of the SWAG system and its operation: (a) Subvines and base module for material loading. As the spool releases the material, the subvine everts and grows along the subvine channels integrated into the sheath. (b) The sheath exhibits a tip unfurling motion, induced by the growth of the subvine.}
\label{fig:Prototyping}
\end{figure}

In this paper, the mechanism was designed with two subvines. As discussed in Section \ref{section:working principle}, this decision was made because the mechanism is intended for dressing assistance on human arms and legs, allowing for the establishment of a primary bending axis. Additionally, a single subvine was not used to prevent unintended jamming caused by asymmetry.

To fabricate the desired structure, TPU-coated ripstop nylon (30 denier, Extremtextil) is cut into a rectangular shape and folded in half, after which an ultrasonic welding machine (VETRON 5064, TYPICAL GmbH) is used to bond the material, forming the cylindrical structure of the subvine. Similarly, the sheath is fabricated by cutting TPU-coated ripstop nylon into a rectangular shape and welding three separate lines to create channels for the two subvines to grow while forming a cylindrical structure for covering objects. Once the subvine is inserted into the end of the sheath’s channel, the basic configuration of a SWAG is completed. When the preloaded material inside the subvine is inflated, the sheath undergoes unfurling, engulfing the target object.

The TPU-coated ripstop nylon exhibits different surface properties on each side. The TPU-coated surface has a higher friction coefficient, which must be considered when arranging the sheath and subvine. The sheath must have strong friction against the clothing material it wraps around or the human body, while minimizing friction against itself. Thus, the TPU-coated side of the sheath faces outward toward the clothing material or the human body. For the subvine, the goal is to minimize friction with the internal materials while effectively restricting relative movement with the sheath and solely transmitting the force at the tip. Therefore, the TPU-coated side of the subvine is positioned outward.

Furthermore, to enable repeated use of a SWAG, the ability to retract the deployed subvine is desirable. Traditional vine robots often face buckling issues during retraction \cite{coad2020retraction}.
However, buckling of the SWAG's subvine is effectively prevented by the structural constraint of the sheath’s channel, enabling a simple yet effective base station design. In this paper, the base was made sufficiently small to pass through the sheath interior, allowing the SWAG to operate without interference in its fully everted initial state.

The base station developed consists of an acrylic cylinder to contain pressure, a reel to store the retracted subvine, and a motor (XL330-M288-T, Dynamixel) to drive the reel. External air is supplied to the acrylic cylinder through an air inlet. Depending on the application, this base station can be customized into different forms. As shown in Fig. \ref{fig:Prototyping}(a), the motor controls the deployment length of the subvine, and upon pressurization, the sheath undergoes unfurling, as shown in Fig. \ref{fig:Prototyping}(b).

\section{Experimental Results}
\subsection{Unfurling force}
\begin{figure}[t]
\centerline{\includegraphics[width=75mm]{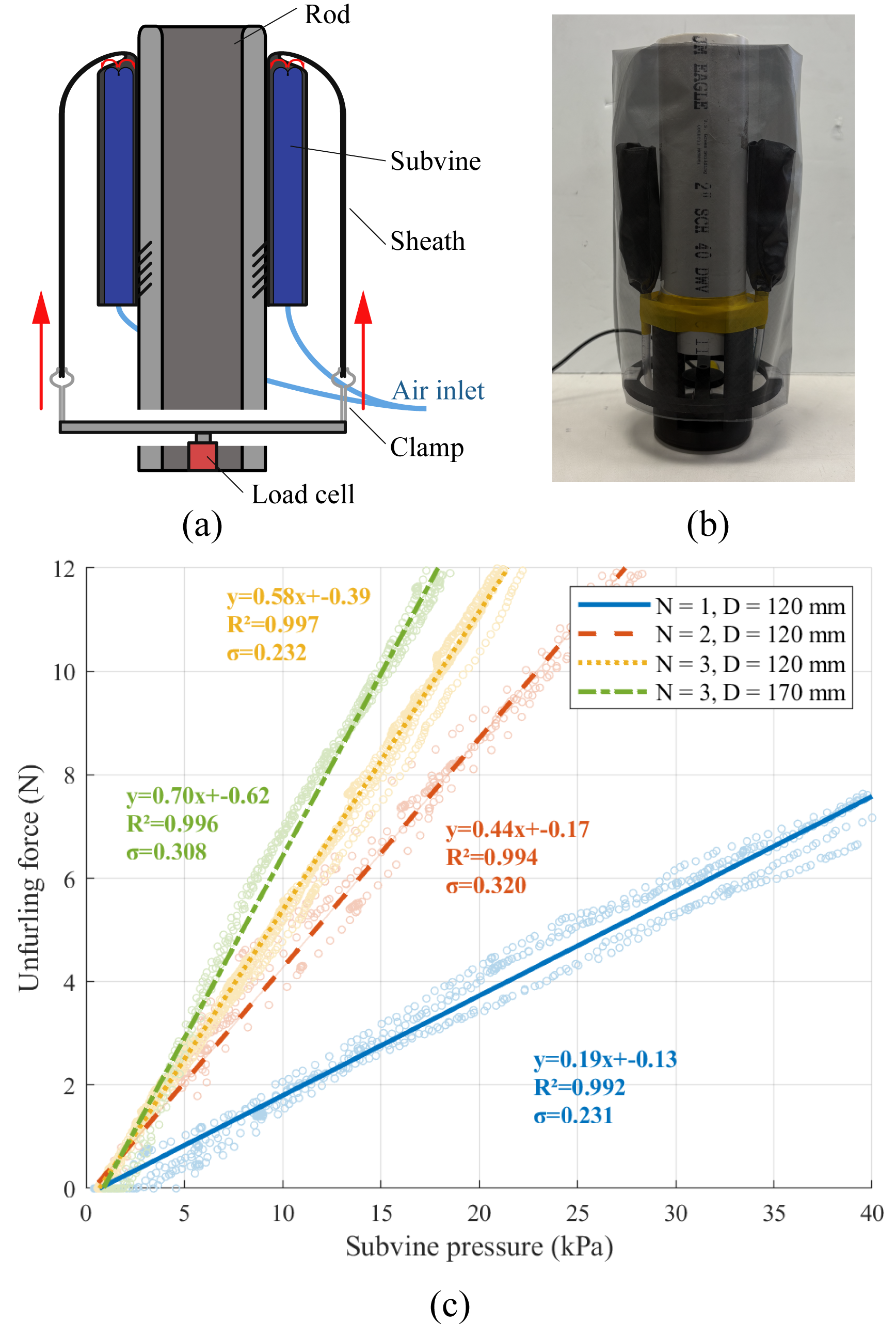}}
\caption{Unfurling force experiment. (a) Schematic illustration of the experimental setup. (b) Actual experimental setup. (c) Unfurling force as a function of subvine pressure for different numbers of subvines.}
\label{fig:EXP1}
\end{figure}

An unfurling force experiment was conducted to quantitatively evaluate the force output of SWAGs, validate the modeling, and characterize the design parameters. Figs. \ref{fig:EXP1}(a) and (b) show the schematic and the actual experimental setup, respectively. Subvines and channels with a diameter of 32 mm, and main tubes with diameters of 120 mm and 170 mm were used for the experiments. To measure the pure unfurling force, the inner sheath, in which the subvine extends, was fixed to a rod during the experiment. As the subvine grew, the outer sheath unfurled into the inner sheath, at which point the force was measured at the end of the outer sheath. The force was recorded using a load cell (UU3, DACELL) directly attached to the end of the outer sheath, while simultaneously measuring the pressure applied to the subvine. Each configuration was tested five times.

Fig. \ref{fig:EXP1}(c) plots the unfurling force against subvine pressure for different numbers of subvines. Each slope was fitted using a first-order polynomial, and the corresponding values of the parameter $a / (a + b)$ for $N = 1$, $2$, and $3$ were calculated as 0.2313, 0.2678, and 0.2841, respectively. Compared to the case of $N=1$, the slope for $N=2$ exhibits a slightly greater than twofold increase, aligning well with the predictions of the model. However, in the case of $N=1$, asymmetric force transmission resulted in a reduction of the measured unfurling force. Additionally, for $N=3$, the expansion of the subvines within the sheath induced jamming between the rod and the outer sheath, leading to a lower force output than predicted by the model. This effect was mitigated by increasing the sheath diameter from 120 mm to 170 mm, confirming that in the absence of jamming, the experimental results aligned with the model predictions. However, an excessively large sheath diameter may lead to positional instability of the subvines during dynamic operation and could result in redundant material when applied to wearable applications. 

As the number of subvines increases, the proportion of space occupied by the subvines inside the sheath also increases, thereby reducing the effective diameter of the donning target relative to the sheath diameter. This effect may limit applicability in most wearable scenarios. Considering these experimental results along with the moment of area analysis conducted earlier, this study adopts a SWAG configuration with $N=2$ to maximize compliance while maintaining sufficient unfurling force generation.

\subsection{Restoring moment and operating pressure in a bent configuration}
\begin{figure}[t]
\centerline{\includegraphics[width=80mm]{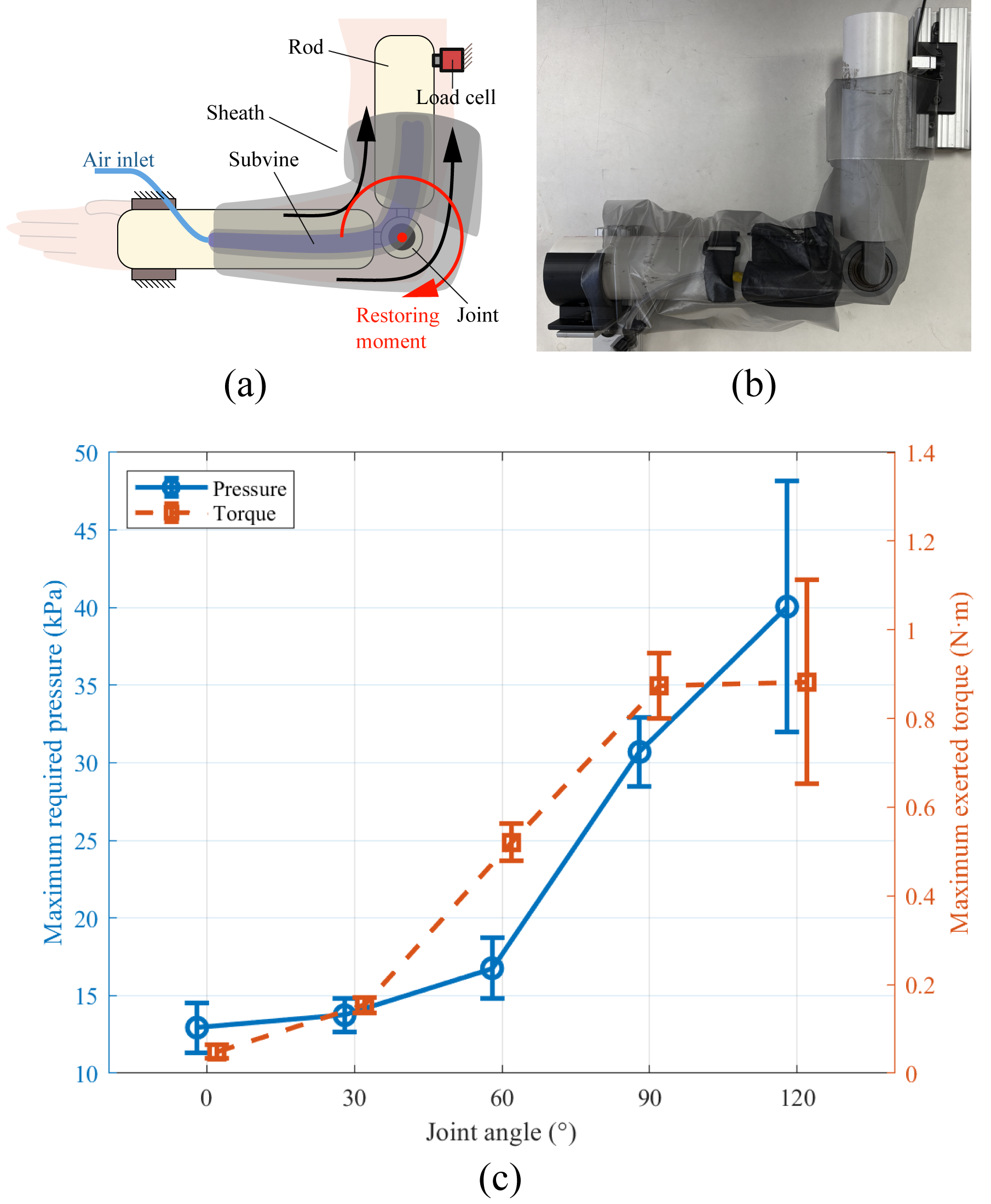}}
\caption{Curvature experiment. (a) Schematic illustration of the experimental setup. (b) Actual experimental setup. (c) Maximum required pressure and exerted torque as functions of joint angle.}
\label{fig:EXP2}
\end{figure}

To characterize the behavior of a SWAG under bending, we investigated its restoring moment and operating pressure through experiments. Figs. \ref{fig:EXP2}(a) and (b) show the schematic and the actual experimental setup, respectively. Since human arms and legs bend at joints, this experiment simulates such a scenario using a joint equipped with bearings and two rods. To measure the pure interaction restoring moment, one rod was fixed in place while the other was left free in the air. When a SWAG was operated, it generated a restoring moment that straightened the suspended rod, causing it to exert force on a load cell grounded to the floor. The applied torque is calculated by multiplying this force by the moment arm from the joint.

Additionally, to characterize the subvine pressure required to pass through each joint angle, the peak pressure required during the unfurling process was recorded for each experimental case. The joint angle was varied from $0^\circ$ to $120^\circ$ in increments of $30^\circ$, and each case was tested five times.

Fig. \ref{fig:EXP2}(c) plots the maximum required pressure and the exerted torque as functions of the joint angle. The measured peak pressure does not necessarily coincide with the moment at its maximum value. Both values generally exhibit an increasing trend as the joint angle increases. The standard deviation also increases with larger joint angles. This phenomenon occurs because, when the sheath experiences sharp bending near the joint, variations in the formation of wrinkles significantly affect the unfurling force required to drag the outer sheath. Although a notable difference in pressure values was observed between the $90^\circ$ and $120^\circ$ joint angles, the resulting torque exhibited only a minimal increase. This is because, at $120^\circ$, the increased pressure was primarily used to overcome the heightened unfurling force due to severe wrinkling, resulting in a transmitted force similar to that in the $90^\circ$ case. Additionally, in both the $90^\circ$ and $120^\circ$ cases, the subvine was observed to be in a collapsed state, which contributed to the similar torque values.

As a result, the SWAG with $N=2$ successfully operated through a $120^\circ$ bent configuration without significant difficulty, and the peak pressure remained within the material's tensile strength limits, ensuring stable operation. Furthermore, the maximum applied torque corresponds to the force required to lift approximately 0.3 kg (300 g) at the fingertips, assuming an average adult male forearm length of 30 cm. This is comparable to the weight of a small beverage can. These results demonstrate the compliance and stable operation of the developed SWAG across various postures.

\subsection{Demonstrations}

\begin{figure*}[t]
    \centering
    \includegraphics[width=145 mm]{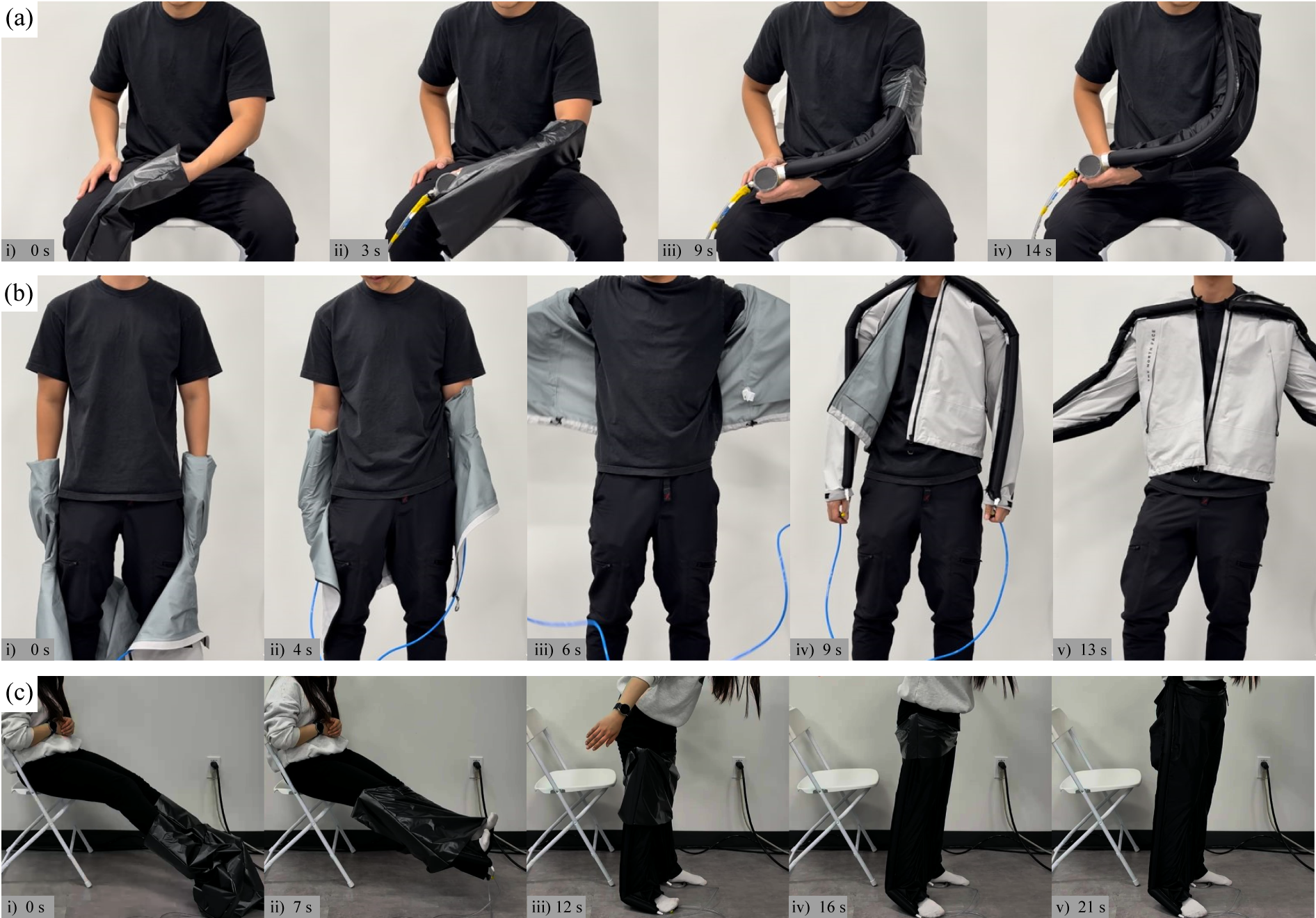}
    \caption{Self-wearing demonstration using the SWAG system. (a) Basic donning operation using a sleeve-type prototype actuated by an automated base station. (b) Jacket donning demonstration with the SWAG integrated as an active component of the garment. (c) Pants donning demonstration with the SWAG integrated as an active component of the garment.}
\end{figure*}

\subsubsection{Basic operation}
To demonstrate the basic operation of the system, we conducted a sleeve dressing trial on a human using the SWAG. For this demonstration, we used the prototype presented in Fig.~\ref{fig:Prototyping}. The base station enables compact material storage and repeatable deployment in the demonstration. The SWAG was actuated under subvine pressure of up to 50 kPa, which proved sufficient to achieve reliable donning performance across various bent-arm postures. The system successfully covered the subject’s arm in 14 seconds, without requiring precise alignment or manual assistance, highlighting its potential for user-friendly operation.

\subsubsection{Jacket and pants donning}
The SWAG can be easily embedded onto conventional clothes, such as jackets and pants. This demonstration employs the preloaded subvine method, as described in Section IV, to validate the system’s compatibility with conventional garment configurations and to showcase its simplified operation without the base station. Jackets and pants are donned within 13 seconds and 21 seconds, respectively, demonstrating the efficiency and practicality of the approach. Jackets with an openable front can be easily donned with subvine growth along simple channels attached on the jacket. The channel paths were deliberately designed to enable effective lifting of the jacket and closure of its front opening. This demonstration shows that, with intentional channel positioning, a SWAG can generate unfurling with directional changes.

\section{Discussion and Conclusion}
This paper introduced the SWAG, an unfurling-based soft robotic dressing system designed to overcome the limitations of traditional robotic technologies in dressing assistance. By leveraging the inherent compliance and adaptability of soft robots, the proposed system facilitates dressing through a growth mechanism, minimizing friction between the garment and the skin. Experimental results demonstrated the effectiveness of SWAGs in enabling safe and adaptive interactions with the user. The demonstrations verify the effectiveness of the proposed system in performing fast and adaptive dressing tasks. The findings of this study highlight several key advantages of the SWAG over conventional robotic solutions. First, its tip-unfurling mechanism eliminates the need for complex manipulation strategies typically required for dressing tasks, thereby simplifying the control framework. Second, its soft and deformable structure enhances user safety, mitigating the risks associated with unintended contact during human-robot interaction. Finally, the system’s rapid deployment capability, intrinsic adaptability that eliminates the need for the user to maintain a fixed posture, and relatively low fabrication cost contribute to its practicality across diverse settings, including home care, medical assistance, and industrial applications.

Beyond its primary function of dressing assistance, the proposed system demonstrates significant potential for broader applications across medical and industrial domains. In medical settings, the system could facilitate wound dressing and assist in covering surgical robots with protective drapes. Moreover, its applicability extends to assisting in dressing pets, humanoid robots, and mobile robots that require protective coverings. The SWAG's simplified structural design and cost-effective manufacturing process further enhance its accessibility, making it suitable for diverse operational environments. Additionally, its capability for rapid deployment is particularly advantageous in emergency scenarios. For instance, the system could assist healthcare professionals and industrial workers in donning protective gear efficiently, aid in the application of compression bandages during medical emergencies, and facilitate the donning of protective equipment in hazardous environments. These attributes highlight the system's versatility and potential to contribute to both routine and critical applications, further expanding the impact of soft robotics in assistive and emergency response technologies.

Further refinements are needed to enhance the system’s functionality and applicability. The current design primarily focuses on dressing assistance for jackets and pants, and future work should explore its extension to more general garments, such as T-shirts and lab gowns, as well as more complex dressing scenarios. While the current system demonstrates reliable performance, incorporating a mechanism that allows the sheath diameter to adjust, or utilizing an elastic material, could further expand its adaptability to a wider range of body types and movement patterns. Enhancing the user interface and automating the garment loading process through the eversion mechanism would further contribute to a more seamless and efficient dressing experience. Given the compact size of the developed mechanism, its integration with robotic arms could further optimize the dressing process. Moreover, evaluating the system’s performance in real-world conditions, including clinical trials with individuals with mobility impairments, will be essential for validating its practical applicability. In addition to donning, the reverse operation of the SWAG’s unfurling process may offer potential for doffing, which could be explored in future research.

In conclusion, the SWAG represents a significant advancement in soft robotic dressing assistance, offering a novel mechanism that prioritizes safety, adaptability, and efficiency. Its versatile design and intuitive operation make it suitable for integration into various assistive technologies, particularly in healthcare and home settings. This system holds promise for advancing assistive robotics and enhancing the quality of life for individuals requiring dressing support.

\bibliographystyle{ieeetr}
\bibliography{main}

\end{document}